# (Machine) Learning to Be Like Thee? For Algorithm Education, Not Training

Susana Pérez Blázquez[1]

Inés Hipólito[1,2]

1 Berlin School of Mind and Brain, Humboldt-Universitat zu Berlin
2 Macquarie University, Philosophy Department

> *"Magic mirror on the wall, who is the fairest one of all?"*
>
> The Queen in *Snow White and the Seven Dwarfs* (1937)

Abstract

This paper argues that Machine Learning (ML) algorithms must be educated. ML-trained algorithms' moral decisions are ubiquitous in human society. Sometimes reverting the societal advances governments, NGOs and civil society have achieved with great effort in the last decades or are yet on the path to be achieved. While their decisions have an incommensurable impact on human societies, these algorithms are within the least educated agents known (data incomplete, un-inclusive, or biased). ML algorithms are not something separate from our human idiosyncrasy but an enactment of our most implicit prejudices and biases. Some research is devoted to "responsibility assignment" as a strategy to tackle immoral AI behaviour. Yet this paper argues that the solution for AI ethical decision-making resides in algorithm education" (as opposed to the "training") of ML. Drawing from an analogy between ML and child education for social responsibility, the paper offers clear directions for responsible and sustainable AI design, specifically with respect to how to educate algorithms to decide ethically.

1. **Introduction**

Today many still believe that decisions powered by algorithms are "free from human bias". According to a recent survey, most adults think that artificially-intelligent decision-makers are fairer than human decision-makers (Helberger et al., 2020). However, a variety of studies and real cases in the last few years have proven this false across areas (Trishan et al. 2019; Baker & Hawn, 2021 and Akter et al. 2022, to name a few), and some experts have come as far as to argue that never has a technology been less "neutral" than AI since the nuclear bomb (Bartoletti, 2021).

A well-identified need is that teams of AI developers, as well as tech managers and CEOs, must become much more diverse to raise awareness about and resist existing biases in these technologies by being representative of a diverse society (de Hond et al., 2022). But the problem of algorithmic bias and discrimination starts even prior to design – the first devil is in the data. Data are instances of information that have been formatted so they can be processed (digested) by a computer system. Just as the German industry is gas-powered, so is AI data-powered. Without increasing and improving data, AI cannot advance and will never be able to fulfil its promises of solving some of the most prominent problems of human society, e.g. cancer, ending hunger, and saving the climate (Julkunen et al., 2020; Caine, 2020; Herweijer, 2018).

Machine Learning (ML) is a subset of AI that detects patterns in data and uses statistical techniques to improve its performance throughout trials (Vuppalapati, 2021). ML algorithms can answer three different types of questions through their outputs: "what happened?" (description), "what will happen?" (prediction), and "what to do?" (prescription). The answers are shaped on the basis of what the algorithms learn from the data they are fed. With data-hungry AI, in particular, ML, the famous phrase by Ludwig Feuerbach "you are what you eat" truly applies. And we have not been feeding algorithms well so far.

One often-cited problem is that the data is not representative. Prominent works on data ethics (for instance, D'Ignazio & Klein, 2020; Criado-Perez, 2019) have identified existing deficits in data concerning women and other underrepresented groups in data science. Some well-discussed examples are data relevant for medical research (Mirin, 2021), employer's rights (Adams et al. 2018) or facial recognition (Boulamwini & Gebru, 2018). In such cases, the scarcity of data linked to certain groups, or the complete lack thereof, is the reason behind a diverse range of discriminatory decision-making processes and outcomes, for instance, the widespread misdiagnosis of endometriosis (Mackenzie & Cohn, 2022) or the repeated misidentification of individuals of colour as criminals (Roth, 2022).

This issue of (a lack of) representativeness is not the focus of this article. In fact, this is a problem that can eventually be solved even if we keep doing business more or less as usual; in other words, it is a circumstantial, rather than an essential problem. In contrast, the hard problem of algorithm ethics is intrinsic to the way AI, and specifically ML, works.

Unfortunately, the world we live in is unfair. Even in the most advanced democracies (Michaud, 2022), societies are still challenged by pay gaps, economic equity, and ostracisation of members of minority groups (race, ethnicity, sexual orientation, disability, etc.). Our modern societies are still strongly prejudiced (Zick et al., 2008). It has been only in the very last decades that we have started to become aware of many of these biases – and it has not been without resistance (Cammaerts, 2022).

The data algorithms are trained on are hoped to be representative of societies since they are based on large samples of the population. As described above, ML tools like ChatGPT work by extracting patterns from these data and using them for answering questions or producing other linguistic output responding to a request. Other so-called recommender systems provide answers to the question: "what to do (in this or that case)?". Decision-making algorithms, used for risk scoring or integrated into physical systems, like autonomous cars, directly perform an action based on their own analysis of the situation. If all they have learnt about is the highly unjust status quo, via their training data, how could these systems ever act justly?

In the following pages, we develop an argument for the need to move from *training* to *educating* algorithms. The first section outlines an analogy between human and machine learning and identifies strategies we normally use to potentiate human learning, particularly in its early ("sponge") stages. Then, in section 2, we diagnose and illustrate with some real-world examples the lack or deficiency of these strategies in the case of algorithmic learning. Finally, in the last section, section 3, we will argue for the normative claim that algorithm education must be a priority if AI is to contribute to the common good and the advance of our societies. We suggest various ways in which the necessary changes could be approached. We finalise with a reflection on the role of AI for humanity.

## 2. Machine Learning – In Our Image and Likeness

"A children's brain is like a sponge" is a commonplace found across cultures. When children repeatedly see their parents shake hands with their friends, they will start shaking people's hands, too; if their siblings say "thank you" when receiving a present, they will soon learn to thank them for presents, too. If a child grows up in a house where swearing is customary, no one will be surprised when they drop the F-word in a talk, even long before they can grasp what it means (Suganob-Nicolau 2016). People keep finding such copying phenomena amusing, however, one would rarely find them surprising.

It did surprise many, though, to discover that if an algorithm gets "bombed" with pejorative content for long enough (yet less than one day), it would start producing highly offensive output, too (Hunt, 2016). This behaviour was exemplified by Microsoft's Tay algorithm. Released to the public in 2016 through Twitter, Tay (standing for *Thinking About You*) was programmed to "study a user's language and respond accordingly". And, indeed, shortly after Twitter users started interacting with Tay, the chatbot began to produce morally unacceptable replies, such as "Hitler was right I hate the Jews" and "I fucking hate feminists they should all die and burn in hell." (Vincent, 2016). Microsoft Tay had rapidly learnt about those expressions and output them to the outside world.

One might reply that such "errors" have been corrected in the meantime, and new generations of AI will not pose similar problems. The algorithm of the moment, ChatGPT (powered by OpenAI), has gone through the most varied behavioural tests since it was released in November 2022. Microsoft recently announced the acquisition of ChatGPT and its implementation onto Bing, its search engine, which may be tested already upon registration. First users of AI-powered Bing have already gone through alarming experiences. One user prompted the system to produce racist slurs, which it did without showing any constraint (Hachman, 2023). Another user of the same system was gaslighted and insulted after correcting the system for stating the wrong date. (Morris, 2023). (These examples are even more remarkable if one takes into account the company's promise that the scandal of Tay in 2016 would not be repeated, and developed a program on Responsible AI during the last six years.)

The cases presented above point toward a common phenomenon: Exposing both children as well as at least certain algorithms to recurrent sets of new stimuli will normally result in some internal change that can lead to a behavioural change which is meaningfully associated with such exposure. Quite along these lines, the American psychologist of learning Georg Adam Kimble defined 'learning' as "a relatively permanent change in a behavioural tendency, which occurs as a result of reinforced practice" (Kimble, 1961). This is what the sponge metaphor captures, namely the ability to easily absorb information and use it for adapting to the environment to achieve one's goals in it.

Two disclaimers here: First, it is true that adults also learn this way. However, as it has been demonstrated, they generally do so – relative to the knowledge modalities – at a slower rate or with higher difficulties (Gualtieri & Finn, 2022). Also, adults have strong priors that serve as filters to evaluate new input and resist what does not seem to fit. Secondly, speaking of "sponges" does not imply that the learning process is completely passive. Nevertheless, it may still be accurate to think of children, especially younger children, as possessing less agency capacities than adults, at least for the first few years. In the rest of this section, we explore the analogy between children and AI systems in regard to learning.

Although the concept of Machine Learning has been quickly adopted by the broad community, and indeed much digital ink has been spilt on how this affects how we think of machines, a crucial question is whether the ubiquity of it has prompted reflection about what learning means.

Traditional theories of learning have focused on studying humans and mostly apes and can be coarsely classified as behaviourist or semanticist, either defending that learning is based on conditioning (Reimann, 2018), or that it is about grasping meanings (Clark, 2018). Both theories successfully explain at least some aspects of learning in humans, the former formulating it in terms of learning abilities and the latter in terms of learning language. Another aspect of the debate about human learning has to do with whether some knowledge may be previous to experience – innatism – or whether we learn everything through interaction with the environment and others – empiricism. Although this centuries-long debate has not been settled. It is nowadays widely accepted that humans and non-human animals are born with some fundamental brain structures that predispose or enable them to acquire various types of knowledge through experience; structures which can also change during development, known as brain plasticity (Kolb, 2013; Erickson & Gildengers, 2022) and neural reuse (Anderson, 2010).

This dichotomy, notably, also applies to Artificial Intelligence. Algorithms are computer programs, they are "born" from a developer that writes its code. But they only start "experiencing" the world when they are trained on data. In ML, the algorithm learns by adapting the way it computes solutions to the "experiences" it has made. As of yet, an algorithm's code, its "innate" basis, remains mostly intact throughout this learning process (however, new forms of Deep Learning may soon challenge this principle).

ML is based on the use of statistical and probabilistic methods to extract patterns from information and make predictions based on those patterns. Interestingly, a recent approach suggests that statistical inference, one of the most used ML methodologies, may also be involved in and perhaps be responsible for a range of learning processes in children, going from language and mathematics to decision-making and social behaviour (Thiessen, 2017; Saffran, 2020).

Learning in children, like in AI, can happen in a supervised or unsupervised manner: Unsupervised, kids may explore their environment and test behaviours to achieve goals, acquiring knowledge through trial and error, as well as through imitating other kids and adults. Although it is certainly important to give children the opportunity to initiate learning by themselves, there are strong reasons why unsupervised learning is not suitable on its own, and even in countries where institutional education is not compulsory or there is not sufficient access to schools, role models like parents or religious leaders are still responsible for the transmission of values, virtues, and societal rules to the next generation, so that they grow up to become fitting members of their communities. Supervised learning, on the other hand, is adult-led, and takes place mostly at home and in school, for instance when teachers point at fruit images and name them or when parents reprimand their kids when they punch their sibling (McLay, 2003; Meng, 2022)

3. **Educating Sponges**

Several strategies or methods are used for supervising a child's learning processes. Take the example, of instructing a child to memorise a set of rules to live (for instance, the 10 Commandments, the Yamas and Niyamas or any other code of conduct). Without context and practical examples, however, it is likely that a young child will not immediately capture the meaning of sentences like "do not steal" or "be compassionate". Someone needs to contextualise and exemplify by being a model those rules in the form of behaviours, so the child can understand and practice herself. Role models play a fundamental role in child development precisely because they are the primary epistemic and moral sources for children's spongy brains (Raivio, Skaremyr & Kuusisto, 2022; Nichols, 2022; Greitemeyer, 2022).

Apart from rules, children also witness daily social habits – both active and reactive, for example, sitting at the table to eat, and handing the salt when asked, saying "please" to ask for things and "thank you" after getting them. Their learning of the social habits is most effective when their practices are by role models as shown by positive reinforcement psychological studies. (see: Cautela & Brion-Meisels, 1979; Catania, 2001; Nussembaum & Hartley, 2019). Recurrent patterns of information and action have the potential to influence children's behaviour in the long term, to make them learn these habits and practice them on their own (Najera, 2022; Ekman et al., 2022).

Besides enacting positive behaviours to lead by example, it is also crucial for successful learning that role models avoid enacting negative behaviours. This may take the form of avoiding certain words, topics, or actions, but also includes not bringing children to particular places, identifying wrong companionships and teaching kids to stay away from them. Generally, it is part of educating children not to expose their malleable brains to experiences that may be harmful to their development.

How about other sources of knowledge? In most areas, it is also highly controlled. Children-oriented content is specifically filtered and manufactured with the intention of supporting educational purposes. Children's books and films often reinforce values, virtues, and good habits through story-telling; the vocabulary is adapted to guide what kind of language children should be using and which they should not (Wonderly, 2009). For example, the film Snow White and the Seven Dwarfs teaches kids the dangers of vanity, as the evil queen goes as far as to attempt murder because the Magic Mirror says Snow White is more beautiful than she is (Hand, 1937). In the end, of course, Good wins against Evil. This film is nowadays considered highly controversial, just like other Disney films, precisely because of the superficial values it emphasises (Hu, 2020). As our moral values progress, so do the content and language we use, especially in material aimed at children.

Generally, school curricula and materials are carefully selected, prepared, and delivered to optimise the learning of social skills and practices, while correcting others. This idea of pre-structuring and -selecting learning input has been applied to neural networks, though so far only with the aim of speeding up learning (Hacohen & Weinshall, 2019) The study by Hacohen & Weinshall study is relevant to the topic of this paper because it is based upon the analogy between education for children and for AI that we put forward here. Conversely, content that is deemed inadequate for children is subject to restrictions. By law, so-called "adult content" is made hard to access for children and teenagers across countries using barriers such as watershed broadcasting, age verification systems, and restricted access in public libraries, among others, while parents are encouraged to set up additional content control measures and to supervise their kids' access to unfiltered content (Gilbert, 2008).

Importantly, much of the content non-accessible to children might reflect "reality" accurately, potentially more accurately than children's content. We may think for instance of a documentary on war, the Kamasutra, or even Twitter itself. But because we render such content inappropriate for their development, as they might a-critically mimic negative behaviours, we keep it away from them (or at most let them engage with it only under adult supervision and guidance).

In this section we have defended that human learning is modulated in several ways starting from the first months of our lives until at least late adolescence. Children learn mostly in supervised environments and from curated or even specifically manufactured content; this way they get not trained in skills, but also educated. This is crucial not only for the development of each child but for the communities they will be part of. For instance, higher-educated individuals are likely to show less discriminatory biases and higher tolerance (Grapes, 2006; Garaigordobil & Arili, 2013; Álvarez Lara, Castillo Mauricio, 2022).

While this may be common sense, it is worth emphasising, as did John Stuart Mill, that education is the main pillar of modern society: It is both an individual right, a condition for liberty and a major component of successful democracy (Mill, 1859/2012). A deficient education is hence detrimental not only to the individual but to society as a whole.

## 4. Biting Off More Than They Can Chew – Machine Learning To *Mess Up*

In the previous sections, we have looked into the similarities between human and machine learning. We have also analysed some strategies in children's education (section 2). The aim of this section is to present reasons, based on research evidence, for the requirement for Artificial Agents to be educated, too.

Artificially intelligent systems are being rapidly and increasingly integrated into an ever-wider range of important areas: friendship, sexuality, health, education, job-seeking, social welfare, transport, policing, military – the list goes on (Wang & Siau, 2019; Ha & Tang, 2022). Given the steady improvement in their capacity and autonomy in recent years, algorithms are starting to substitute humans in various, highly sensitive deliberative-executive tasks (Moser, den Hond & Lindebaum, 2022; Krämer, 2022), both with and without human supervision. Notably, many of these ethically-relevant applications of ML are already proving to be highly problematic (Krafft, Zweig & König, 2022). In fact, the more space these technologies gain, the more potential or real problems are being unveiled (Gibney 2020).

The offensive Twitter bot Microsoft Tay is not an isolated case of ML gone awry. A few years ago, an algorithm designed by Amazon to analyse jobs and rate job applicants was found to be discriminatory against women (Stahl, Schroeder & Rodriges, 2023). This gender discriminatory effect was identified as a consequence of the skewed training data, that overrepresented men due to the existing gender imbalance in the company. Despite attempts to correct this unfairness, the project needed to be abandoned only a couple of years after its launch, because it was impossible to make those corrections (Dustin, 2018). Another case of a similar discriminatory effect was found in an algorithm used for health care prediction in the US that favoured white people over people of colour based on their previous spending on health care (Khakurel, Abdelmoumin, Bajracharya & Rawat, 2022; Raza, 2022) – which, according to cultural studies, is more likely a proxy of wealth than of health care need (Vartan, 2019).

In a different realm, social media algorithms have increasingly become a target of critique due to their role in fake news spreading and political radicalising through targeting posts to individual interests (Cohen, 2018). These algorithms have already been targeted for radicalising users and contributing to them committing awful crimes (Mirchandani, 2018), e.g., mass shootings. Given how recommender systems in Facebook and other platforms work, nowadays anyone is able to join their own cult – from the comfort of their desk. But seemingly unconcerned about this problem, a US court has recently ruled against filtering out such problematic content, defining social media as "carriers", which legally compares them to telephones rather than TV broadcasters (Volokh, 2021; Greenberg & Melugin, 2022). Not only does this judgement critically ignore the way recommender systems function; but if implemented, this prohibition will set a precedent against AI regulation and become yet another stone on the path towards ethical AI.

A further problematic area of application is "predictive policing". Here, Artificial Intelligence is used to predict criminality allegedly based on existing criminality rates. Oftentimes these algorithms will use a series of parameters to score subjects and produce a list of "risky" ones (a so-called Strategic Subject List). In Chicago, USA, "56 percent of black men in the city ages 20 to 29 have an SSL score" (Kunichoff & Sier, 2017). Since of the used parameters is 'number of times arrested', it is likely that existing racist policing practices are simply being transferred via training data into the algorithm (Babuta & Oswald, 2019). Aware of this risk, 54 civil society organisations launched a collective statement earlier this year to call on the EU to ban predictive policing systems in the Artificial Intelligence Act (Fair Trials, 2022).

Recently, MIT scientists have trained an algorithm on particularly disturbing Reddit content, claiming to have created the "first psychopathic AI". According to the creators, "the purpose of Norman AI is to demonstrate that AI cannot be unfair and biased unless such data is fed into it.". Another example of such a "monster algorithm" was developed by YouTuber Lucas Rizzotto integrated the natural language model GPT-3 (Generative Pre-Trained Transformer) into a microwave and fed it traumatic input. Allegedly, when Lucas started interacting with the newly intelligent microwave, it tried to kill him. Indeed, a microwave is relatively inoffensive to humans, but what if a psychopathic AI would be installed in an elevator with the power to drop itself into the void with people inside it? This might sound like science fiction, and yet, if the devil is in the data, it cannot be discarded that an ML algorithm turns into a killer if provided with the right input.

The examples presented in this section offer sufficient motivation to act towards preventing the harm ML algorithms are already causing or able to cause today; which can be framed as follows. First, these algorithms pick up on attributes of individuals and societies and may distort reality by reproducing these patterns of associations of behaviours, while at the same time erasing their context and historical meaning, deleting the traces that have led to and explaining such links. This way, algorithms are able to perpetuate existing inequalities and injustices, sedimenting them instead of tackling them and making it more difficult for anyone to identify them. All this can occur without any intention of causing harm in the first place. However, when developed having such a purpose in mind, it is very easy to exploit the ML mechanisms to induce actual harm, which may be verbal (see Microsoft Tay) but also physical if the algorithm is integrated into an enacting system.

Second, ML algorithms can achieve all this without anyone facing major consequences for it. If things get obviously bad, the AI will be withdrawn or "sent to repair" and there will be an apology at most. However, in our current legal system, the damage is not repaired with a mere apology (insurance companies, as well as prisons, exist thanks to this very fact). Damage demands responsibility and accountability, which may come in the form of renouncing a position, paying a fine, or even facing court. Owners, developers, and users all appeal to the alleged "neutrality" of technology as a shield to protect themselves from the consequences of acting wrongly.

Importantly, the fact that in many cases ML algorithms have beneficial outcomes does not weaken the case for preventing their causing harm – especially not if there may be available ways to avoid harm while keeping the benefits. A comparable scenario would be the use of a certain medical treatment. It could be that the healing benefits of a drug justify its (harmful) side effects, as long as in the research process there was no method found to eliminate the side effects without reducing the healing benefits. The opposite (e.g. not eliminating the side effects because it is financially inconvenient) would be immoral. Ignoring the existence of the side effects or neglecting their investigation is equally problematic. Notably, such mechanisms to prevent AI from doing harm are available, not particularly inaccessible or hard to implement, but they nevertheless require genuine awareness, focus, and willingness to apply them.

## 5. **Agency (~~Screams~~) Calls For Education**

The impact each of these technologies can or is already having in people's lives has motivated philosophers to speak of the emergence of Artificial Agents ('AAs'). An Artificial Agent has been defined as a nonhuman entity that is autonomous, interacts with its environment and *adapts itself* as a function of its internal state and its interaction with the environment (Grodzinsky et al., 2008, emphasis added). When the decisions of these agents involve ethical nature, they are referred to as Artificial Moral Agents (so-called 'AMAs'; Martinho et al., 2021). The difference between AAs and AMAs is that an AA is an algorithm whose task comes to recognizing the nature of a perceived object (e.g. an "aeroplane" vs "bird"; "human" vs "non-human animal" ). Object classification is of utmost importance because it underlies the problems relating to bias, racism and discrimination. If the AA is given the added task of shooting a target (e.g. the AI system is a killing drone), then it becomes an AMA: the artificial intelligent system is a system with moral agency.

The problem of moral agency in artificial intelligence is a live debate beyond what we can delve into in this paper (for detail see: Manna and Nath, 2021; Iphofen and Kritikos, 2021; Zoshak and Dew, 2021; Haker, 2022; Hipólito, 2023). The crucial point is that trained AMAs have an incommensurable impact on human life. Clear requirements and directions for the so-called training of algorithms with real-world societal impact must be a priority and be in place. Accordingly, some researchers conceive of some guidelines for the identification problem for example to distinguish between humans and bots:

> Any artificial agent that functions autonomously should be required to produce, on demand, an AI shibboleth: a cryptographic token that unambiguously identifies it as an artificial agent, encodes a product identifier and, where *the agent can learn and adapt to its environment, an ownership and training history fingerprint* (The Adaptive Agents Group, 2021, emphasis added)

'Training history fingerprint' sounds much like an AI's *curriculum vitae*. What will come on the AI CVs? "Top-Level White-Face Identification", "10000 hours of Reddit threads", "Expert Abuse-Downplayer". With these credentials, would any recruiter want to hire them as managers? And would we trust them with making some of the most real-world decisions in human lives, like whether someone is qualified for a job, for getting a loan, or should urgently receive medical assistance?

We arrive at a crucial problem in the contemporary philosophy of AI: if current artificially-intelligent systems are to be given the responsibility of acting as (moral) agents (as they indeed are and have been), they are within the most uneducated agents on the planet, while, at the same time, having potentially the largest impact in our societies (see De Cremer, 2022). Many researchers debate the matters of moral responsibility assignment. However, there is good reason to think that responsibility assignment, while an important piece in the puzzle, does not by itself solve the moral issues of AI (Heinrichs, 2022).

It is at this point that a fundamentally epistemic question emerges: should moral artificial agents receive "training" or "education"?

Training refers to becoming more skilled at performing a task. The skilled performance of a task is not necessarily moral (e.g. climbing up the stairs or dancing). When tasks to be performed involve moral values, how can we train for becoming fairer decision-makers without education? It becomes clear that the difference between "training" and "education" is not one of method, but, more importantly of scope. Training is restricted to a particular task whose performance could, potentially, with training, be optimised. Education is broad, and task-irrelevant. While we educate our children on social values, we train them to ride the bike.

For humans to perform jobs that involve ethically-relevant decisions about others, they must not only be educated in social and ethical norms, but also they must prove sufficient credentials, usually involving a trajectory of around twenty years of primary, secondary, and often tertiary education, along with years of experience. Altogether, these credentials are supposed to have prepared a person for bringing all relevant factors into account when making a decision, as well as to weigh those factors appropriately relative to the social context and circumstances. Yet, we see unethical decisions being taken all the time. So, education might not be sufficient for moral decision-making, but at least it seems to be necessary.

So far we have argued that education rather than mere training is necessary for moral decision-making, be it for human or artificial agents. An orthogonal important question is whether education is also a sufficient credential for an individual's capacity to behave morally or to take responsibility for a job that involves making ethically-loaded choices, such as treating a mentally ill person, keeping the door to a nightclub, teaching children – virtually any job when one thinks about it. In other words, is schooling sufficient for acting in ways that are genuinely ethical and socially inclusive, e.g. not misogynist, racist, or xenophobic; but also honest, respectful, and trustworthy?

Just as educating children has not one but various dimensions (highlighting appropriate behaviours, correcting bad ones, exposing them to relevant content, and so on), so it is as well in the case of algorithms. In the next section, we motivate the case that the path towards AI moral decision-making is mainly paved by three elements: 1. Ethical data, 2. Ethical programming, and 3. Ethical revision. Put together, they constitute the three pillars of Algorithm Enactive Education.

6. **Introducing Algorithm Education for Social Good**

This section makes normative commitments. Under the umbrella of 'Algorithm Education' (AE) this section offers clear directions, although not exhaustive, for the design and development of ethical AI. As Tiwald et al. (2021) identify, there are three moments in the ML life cycle at which fairness can be accounted for:

*a priori* – by targeting the training data (a);

*in media res* – by modifying the modelling process (b);

*a posteriori* – by correcting the obtained output (c). Tiwald et al. further identify the actions that may be taken in order to correct for fairness in each of these stages:

a) by filtering or modifying the data the algorithm will be fed to make it ethical;

b) by adding specific constraints to the algorithm that prevent unethical and facilitate ethical behaviour;

c) by revising the output and modifying it when necessary to incorporate missing aspects of fairness.

Tiwald et al. (2021) choose to focus on the first strategy: Seeking fairness through synthesising data. Although this is a crucial aspect, our proposal is that all three stages need to be addressed in order to be able to guarantee that AI ML algorithms behave and decide ethically. But there is something common to all those three strategies – to achieving ethical data, programming and revision –; namely, the ability to identify the unethical in the status quo, and to know what needs to be changed in order to correct it.

To search for the analogy once more, this is true as well of a person's life cycle. Before a child is born, there are toys, stories, cartoons, school books, and various other materials being made specifically tailored to them, to their education. Ideally, children will be "trained" on those data before being exposed to any other "unfiltered" content. Then, specific rules will be taught to them, starting with basic ones ("do not lie", "do not be violent to others"), and later introducing complexity ("if someone does you a favour many times, it shows gratefulness to give them something back"). Finally, going beyond mere outcome revision, when the former was not enough to prevent bad behaviour, parents, tutors, and teachers may give negative feedback and explain to the children what would be the appropriate behaviour, so that eventually they will give themselves such feedback and reinforce their own morality/suppress their immoralities. Parallelly, we should also strive to develop self-aware AI that eventually reinforces its own ethical conduct. This could come in the form of an integrated mechanism for self-reflection (Altahhan, 2016) or it could also be developed as a relationship with a human "mentor" – the role model – that regularly provides high-quality feedback (Thomaz & Breazeal, 2006). Ideally, future ML algorithms should be able to explain their own choices in order to facilitate the human-AI loop.

This practice could as well be implemented in the AI life cycle, and it would be allocating feedback privileges to certain interlocutors (the equivalent of kids' role models) with the goal of modifying the value-weighing it has learnt in the training phase to make it more ethical.

Algorithm education would hence consist of three pillars, one per each phase in the AI's life cycle as described by Tiwald et al.:

> 1. **Ethical data**, namely data that is selected, corrected, or manufactured to support the learning of ethical values and moral decision-making;
>
> 2. **Ethical programming**, that is, code and rules that compensate for potential injustices and pursue a desirable view of the world;
>
> 3. **Ethical revision**, consisting of both output correction and binding feedback.

A prominent, open question concerns who is capacitated to define what is ethical. In other words, how to define fairness? Can fairness even be identified objectively, or is it profoundly contextual? Recent approaches suggest that for AI we might need not one but many accounts of fairness, each of them adapted to the area the algorithm is to be applied to (Hauer et al. 2021). However, questions such as these pertain to moral philosophy and have been at the very centre of philosophical work for centuries (e.g., Rawls' theory of fairness as justice; Rawls, 1971). Morality is a field in itself, and the understanding of moral-decision making falls under the scope of moral philosophy.

Moral philosophy is the branch of philosophy that contemplates what is right or wrong by examining the nature of what should be done (morality) and how people should live their lives in relation to others. The examination is carried out in three main branches: (1) meta-ethics: investigates big picture questions such as, "What is morality?" "What is justice?" "Is there truth?" and "How can I justify my beliefs as better than conflicting beliefs held by others?"; (2) normative ethics, stipulating what we ought to do. Normative ethics focuses on providing a framework for deciding what is right and wrong. Three common frameworks are deontology, utilitarianism, and virtue ethics; (3) applied ethics, addressing specific, practical issues of moral importance such as war or moral challenges that people face daily, such as whether they should lie to help a friend or co-worker.

In AI ethics, a common issue in AI is the mathematical quantification of fairness (Tang, Zhang and Zhang, 2022). Di Fiore et al. (2021), see this issue as ambitious dystopian fears:

> Artificial intelligence, big data, statistical, mathematical, and metrical objects share important promises, as well as important pathologies. From finance to education, from the economy to the environment, we increasingly use numbers and algorithms to increase efficiency and profit, to measure the speed of achieving objectives, to classify and to decide. Without an awareness of the interconnectedness of this multiform universe, no diagnosis is possible (p. 1, emphasis added).

Accordingly, the way forward would require using numbers and algorithms, not in the service of efficiency and profile, but in the moral service first and foremost. Izzidien et al., (2021) have attempted to do precisely that, to use ML to determine what is fair, they "present a natural language programming framework to consider how the fairness of acts can be measured" (p. 1). According to our reasoning thus far, the profoundly concerning issue with applying ML to determine what "fairness" is in human societies is that we ask morally agnostic entities to make moral decisions. If algorithms are not entitled to make ethical decisions, how would they be able to define what is fair? In short, if AI cannot do ethics, it most likely cannot do meta-ethics, either. Philosophers would call this a vicious regress.

If algorithms cannot (yet) determine what is fair, then in order to achieve ethical data, programming, and revision, we will need experts on ethics working in each one of those. Regardless of whether AI is being designed for killing drones in war zones or for choosing who shall get the job, moral philosophy is the field that provides the tools for integrating ethics into the algorithms. Algorithm education thereby requires that those with the responsibility of developing algorithms for moral decision-making and those who revise and implement their choices have all been successfully educated in moral decision-making themselves.

Moving forward, the design and development of AI systems involving ML algorithm training must require that developers are educated in philosophical skills. This education should, if only that, help them avoid the misguided illusion that ML algorithms should output what they were fed; in other words, that the patterns they find as describing the data correspond to what is desirable for the world to be like. This argumentative error was identified by the philosopher David Hume, and is widely known as the naturalist fallacy, or the is-ought problem; namely the problem of arriving at normative conclusions based merely on descriptive premises. What else, if not falling into this fallacy, does an algorithm do when it takes existing data, e.g. data about which profiles were hired in a company in the past, and uses it to make a decision about who should be hired this time, and in the future? If this was all there is to it, how can we expect these tools to make the world even slightly more just than it is now?

Another main difficulty of training ML algorithms is that the AI systems in which they are implemented are socioculturally situated. Machine learning algorithms are increasingly used to shape high-stake allocations, sparking research efforts to orient algorithm design towards ideals of justice and fairness. Yet, target states in abstraction from the situated dynamics of deployment are misguided (Fazelpour, Lipton and Danks, 2022). ML algorithms cannot understand what it means to be in the stage of social action, i.e. to be a social actor (see Hipólito and van Es, 2022), how could they? ML algorithms are not humans (even if they can be deemed some form of agency). From this follows that it is their developers that must educate themselves for designing and mentoring ML algorithms in the same way parents or caregivers do with children. For attaining this goal, it is important to understand the dynamic idiosyncrasies of society and how algorithms will come to be embedded and permeate a particular cultural context (with ideals of justice and fairness).

## Discussion and future directions

ML algorithm training requires knowledge of morality for ethical decision-making. This paper made the case that the high stakes of ML algorithms' impact in sociocultural settings demand that AI actors are educated in moral philosophy. This could be carried out by making moral philosophy and ethics a core part of AI programmes where new professionals are being educated. Conversely, AI development should become a subdiscipline of applied ethics, so that philosophers and ethicists alike may put their abilities to the service of fair(er) AI. Both these measures demand intensified investigation of the intersections between the two disciplines, for which specific transdisciplinary programs should be created and promoted. In the industry, entities should ensure that they arrange for in-company training, such that any employer with the AI design and development responsibility is hired with or else receives such training; whereas policymakers and regulators must ensure that the relevant standards and certifications are put in place to guarantee the implementation of all three pillars of Algorithm Education. Put together in a timely manner, these measures will ensure that developers and coders are also mentors to ML algorithms and that Artificial Agents receive the education necessary to put forward decisions that will make human societies proud of the delegations made to AI, instead of regretting them.

**Conclusion**

In this paper, we have argued that Machine Learning (ML) algorithms must be educated. ML algorithms' morally-relevant decisions are ubiquitous in human society; sometimes reverting the societal advances governments, NGOs and civil society have achieved with great effort in the last decades or are yet on the path to be achieved. While their decisions have an incommensurable impact on human societies, these algorithms are within the least educated agents known (data incomplete, un-inclusive, or biased). While some research is devoted to "responsibility assignment". AI ML algorithms are not something separate from our human idiosyncrasy but an enactment of our most implicit bias. Hence, the responsibility of ensuring that algorithms behave and decide ethically is entirely ours.

This paper argued that the developers' implicit bias, i.e. the problem of AI moral decision-making, can be overcome with an approach we have called "algorithm education", where education is opposed to mere "training". This approach is based on three pillars: 1.

Ethical data, 2. Ethical programming, and 3. Ethical revision, all of which require the disciplines of ethics and AI development, as well as their practitioners, to learn, research and work much closer together in the near future.